\documentclass{sig-alternate}

\makeatletter
\newif\if@restonecol
\makeatother

\usepackage[ruled, english]{algorithm2e}

\newcommand{\diag}{\mathop{\mathrm{diag}}}

\renewcommand{\vec}{\mathop{\mbox{vec}}}

\usepackage{mathtools}
\usepackage{amsmath}
\begin{document}

\long\def\/*#1*/{}
%
% --- Author Metadata here ---
\conferenceinfo{21st ACM SIGKDD'15}{Sydney}
\CopyrightYear{2014} % Allows default copyright year (20XX) to be over-ridden - IF NEED BE.
%\crdata{0-12345-67-8/90/01}  % Allows default copyright data (0-89791-88-6/97/05) to be over-ridden - IF NEED BE.
% --- End of Author Metadata ---

\title{Tensor SimRank for Heterogeneous Information Networks}


\numberofauthors{2} %  in this sample file, there are a *total*
\author{
% 1st. author
\alignauthor
Ben Usman \\
       \affaddr{Skolkovo Institute of Science and Technology, Novaya St., 100,  Skolkovo,
143025, Russian Federation \\ Moscow Institute of Physics and Technology, Institutskiy Lane 9, Dolgoprudny, Moscow, 141700, Russian Federation}\\
       \email{ben.usman@skoltech.ru}
% 2nd. author
\alignauthor
Ivan Oseledets \\
       \affaddr{Skolkovo Institute of Science and Technology, Novaya St., 100,  Skolkovo,
143025, Russian Federation \\ Institute of Numerical Mathematics, Russian Academy of Sciences, Gubkina St., 8, Moscow, 119333}\\
       \email{i.oseledets@skoltech.ru}
}

\date{1 Feb 2015}
% Just remember to make sure that the TOTAL number of authors
% is the number that will appear on the first page PLUS the
% number that will appear in the \additionalauthors section.

\maketitle
\begin{abstract}
We propose a generalization of SimRank similarity measure for heterogeneous information networks. 
Given the information network, the intraclass similarity score $s(a, b)$ is high if the set of objects that are related with $a$ and the 
set of objects that are related with $b$ are pair-wise similar according to all imposed relations.  
\end{abstract}

\category{}{Information systems}{Retrieval models and ranking} -- \textit{Similarity measures}

\terms{SimRank, Probabilistic SVD, Tensor, Low-rank approximation}

\section{Introduction}
Most data in the modern world can be treated as an information network, thus network node similarity measuring has wide range of applications: search \cite{page1999pagerank}, recommendation systems \cite{konstan1997grouplens}, research publication networks analysis \cite{giles2006future}, biology \cite{roy2007integrative}, transportation and logistics \cite{jiang2005knowledge} and others.

Consider a semantic network: set of types $\mathcal T$, each type $t \in \mathcal T$ is a set of entities; set of relations $\mathcal R$, each relation is 2-order predicate defined on two types from $\mathcal T$: 
$${\mathcal R \ni r_{t p} : t \times p \mapsto \{ 1, 0 \}}, t, p \in \mathcal T,$$
 both types in relation can be equal ($r_{tt}: t \times t \to \{0, 1\}$), few relations can share the same pair of types ($\exists r^{(1)}_{tp} \neq r^{(2)}_{tp} \in \{0, 1\}^{t \times p}$). That structure may be considered as a graph with colored vertices and colored edges: vertex color is its entity type, edge color correponds to a relation. 

%\begin{figure}
%\centering
%\includegraphics[width=0.5\textwidth]{pic1.png}
%\caption{Heterogenious network as a colored graph}
%\end{figure}

The question that we address is how to define similarity functions 
$$s_t: t \times t \to \mathbb R , \quad \forall t \in \mathcal T,$$
that would reflect the closeness of objects based on "similarity of relations" they enter, and at the same time not mixing different relations as soon as "objects of different types and links carry different semantic meanings, and it does not make sense to mix them to measure the similarity without distinguishing their semantics" \cite{lee2013pathrank}.

\subsection{Related work}
 The basic graph structure similarity measure is the classical SimRank \cite{jeh2002simrank} over a homogeneous graph $G = (V, E)$ which is defined as follows: 
 $$N_G(a) = \{v \in V : (v, a) \in E(G) \},$$
 $$ s(a, b) = \frac {C}{I(a)I(b)} \sum_{\substack{v \in N(a) \\ w \in N(b)}} s(w, v).$$
The main drawback of this approach is that we cannot induce multiple relations or object types, so the only option is mixing them up into blobs "relation exists" and "all objects" that is completely not applicable in the case we have multiple relations with different semantics, for example the OpenCyc ontology node of the concept "Game" (see Figure ~\ref{fig:opencyc}) cannot be easily expressed via a single type of relations and objects.

\begin{figure}
\centering
\noindent\fbox{%
    \parbox{8.2cm}{%

\textbf{A Type of}: devised structured activity

\textbf{Instance of}: candidate KB completeness node, clarifying collection type, type of object, type of temporally stuff-like thing

\textbf{Subtypes}: board game, brand name game, card game, child's game, coin-operated game, dice game, electronic game, fantasy sports, game for two or more people, game of chance, guessing game, memory game, non-competitive game, non-team game, outdoor game, party game, puzzle game, role-playing game, sports game, table game, trivia game, word game

\textbf{Instances}: ducks and drakes, ultimate frisbee, darts, pachinko, Crossword Puzzle Activity CW, pool, Snooker, mini golf

    }%
}
\caption{OpenCyc ontology node of concept "Game"}
\label{fig:opencyc}
\end{figure}

Personalized PageRank \cite{jeh2003scaling} is also often used to measure similarity in homogeneous graphs:
$$ \pi_a(b) = \varepsilon \delta_a(b) + (1-\varepsilon)\sum_{(w, b) \in E} \frac {\pi_a(w)}{\alpha_{w, v}} , $$
that it same as PageRank, except random jumps are made into some pre-chosen node $b$, rather then into random node.
 
Another option is PathRank \cite{lee2013pathrank} that measures path-similarity between objects $a, b$ picked from the same class $A$ of the heterogeneous information network $\mathcal N$ given a symmetric meta-path $\mathcal P$ (set of paths that satisfy composition of relations $M_{1}, M_2, \dots , M_n$ that $A \xrightarrow{M_1} C_1 \xrightarrow{M_2} C_2 \dots \xrightarrow{M_n} A$, so $A \xrightarrow{M_1 \circ M_2 \circ \dots \circ M_n} A$) as a number of paths from the object $a$ to the object $b$ (each step $i$ must satisfy corresponding relation $M_i$ in $\mathcal P$) normed over the number of paths from $a$ to $a$ plus the number of paths from $b$ to $b$ given $\mathcal P$:
$$s_{\mathcal P}(a, b) = \frac {\| \{ p \in \mathcal P: a \xRightarrow{p} b \}\|}{{\| \{ p \in \mathcal P: a \xRightarrow{p} a \}\|} + {\| \{ p \in \mathcal P: b \xRightarrow{p} b \}\|}}$$ 
That approach can handle several relations and object types and is very useful when we know the structure of relations we want our similarity measure to be based on. In case we want just to "put our relations into a black box" that would find similarity that would capture all 
network relations as a whole, we might want to use something different. Recently, an approach \cite{sun2013mining} for building an optimal linear combination of meta-paths has been proposed.

There are several works on measuring similarity between objects from different classes, see, for example,  \cite{shi2014hetesim}.
\section{Tensor SimRank}
\subsection {Problem statement}
Let us consider a function $s_t(a, b)$ that assigns similarity score for two objects from the same class $t$ as follows: objects $a, b \in t \in \mathcal T$ are similar (value $s_t(a, b)$ is high) if they relate to objects which are similar too.
That interdependence can be expressed via the following definition:

\begin{equation*}
    \begin{split}
        &N_{r_{t p}} (a) = \{ b \in p \vert r_{t p}(a, b) = 1 \},\\
        &s_t(a, b) = \frac 1 Z \sum_{ r_{tp} \in \mathcal R } w(r_{tp}) \sum_{ \substack{c \in N_{p}(a) \\ d \in N_{p}(b)} } s_p(c, d),
\end{split}
\end{equation*}
where $r_{t p}$ is the relation between classes $t, p \in \mathcal T$, $N_{r_{t p}}$ is the neighbourhood function that returns set of objects from the class $p$ that are related to the object $a$ via the relation $r_{t p}$, $w(r_{tp})$ are the weights corresponding to the relation $r_{tp}$, $Z$ is the normalization constant.

This can be rewritten as a \emph{Tensor SimRank equation}:
\begin{equation} \label{eq:tensor_simrank}
    \begin{split}
        &s_{\alpha\beta} = \sum_{\gamma} w_{\alpha\beta\gamma} \ \textbf r_{\alpha\beta\gamma} \ s_{\alpha\beta} \ \textbf r_{\beta\alpha\gamma}, \\ 
        &s = \text{diag}( \{ s_{t} \}_{t \in \mathcal T}) , \quad s_{\alpha\alpha} = 1,
\end{split}
\end{equation}
where $s$ is a block-diagonal matrix (one block per each entity type), $w$ are the relation weights, $\textbf r_{\alpha\beta\gamma}$ are the \emph{stochastic relation tensors} \footnote{We have to use tensors instead of matrices to have multiple relations on the same pair of classes} (which have non-zero blocks where relations exist). 
%That can be flattened into 
%\begin{equation} 
%\label{eq:tensor_simrank_flat}
%s = w \cdot \textbf r \odot_2 s \odot_1 \textbf r^T ,
%\end{equation}
%$$ \textbf r \odot_2 s 
%\stackrel{\mathclap{\normalfont\mbox{def}}}{=} r_{(3)} 
%\cdot I_n \otimes s , \quad \textbf t \odot_1 \textbf r 
%\stackrel{\mathclap{\normalfont\mbox{def}}}{=}
%t_{(1)} \cdot (r_{(2)})^T$$  

%n is total number of objects and $r_{(i)}$ is the i-th mode   of $\textbf r$.

Similarity scores between elements of different classes are equal to zero by the definition.
Relation between objects of unrelated classes is equal to zero by definition too. 
Equation \eqref{eq:tensor_simrank} is basically the classical SimRank equation with the adjacency tensor instead of the adjacency matrix: each non-zero layer of tensor encodes some relation on the same pair of types. If one has more than a single relation between types $p, t \in \mathcal T$, then $\textbf r$ would have multiple non-zero layers on the intersection of indices associated with the classes $t, p$ --- one adjacency matrix per layer. In \eqref{eq:tensor_simrank} the index $\gamma$ stands for (weighted) summation over all layers of the tensor. 
That can be equivalently rewritten explicitly:
\begin{equation} 
\label{eq:tansor_simrank_matrix_sum}
S = \sum_{\gamma} w_{\gamma} W_{\gamma} S W_{\gamma}^T + D, 
\end{equation}
where the diagonal matrix $D$ has to be chosen in a such way that $\diag(S) = I$.
\subsection{Computational algorithm}
Simple iterations for \eqref{eq:tensor_simrank} are computationally demanding due to large-scale matrix-by-matrix products, thus we propose a a method that exploits the fact that $s$ is block diagonal and $\textbf r$ is a three-dimensional block tensor with size of the last dimension (number of layers) much less then the overall amount of objects. On each iteration $k$ for each $r \in \mathcal R$ we recompute $s_i$ updates independently (assuming all other $s_j$ fixed), see Algorithm~\ref{tensorsimrank:alg1}.

\begin{algorithm}[h]
    \caption{Idea under Tensor SimRank}\label{tensorsimrank:alg1}
 \KwData{$\mathcal T$ - classes, $\mathcal R$ - relations}
 \KwResult{ $\mathcal S = \{ s_t(a, b) \}_{t \in T}$ }
 \Repeat{$ \sum_{t} \| s_t - s_t^{next} \| < \delta$}{
  \For{$s_t \in \mathcal S$}
  {
    assume all $S \setminus s_t$ fixed \\
            \For{$r \in \mathcal R : r_{tp}: t \times p \mapsto \{ 1, 0 \}$}
        {
    \For{$(a, b) \in t$}
    {
            \For{$(c, d) \in p$}
            { 
\begin{align*} 
s_t^{next}(a, b) & \mathrel{+}= r_{tp}(a, c) s_{p}(c, d) r_{tp}(b, d) \\ 
				 & \mathrel{+}= r_{tp}(a, c) s_{p}(c, d) r_{pt}(d, b) 
\end{align*} 
            }
        }
    }
  }
  update all $s_t \leftarrow s_t^{next}$
 }
\end{algorithm}
So we just update the similarity score for each class assuming all other classes similarities are fixed in a way that the objects from the target class ($t$) that are related to objects from some other class $(c, d) \in p$ that are close ($s_{p}(c, d)$ is high) become closer too ($s_{t}(a, b) \uparrow)$.

To show actual vectorized algorithm of similarity computation, let us introduce some additional notations: set of entity types $\mathcal T = \{ t_i \}_{i=0}^{N}$, each entity type $t$ is a set of entities, set of symmetric relation functions $\mathcal R = \{ r_{tp}^{(j)} \}_{j=0}^L$ where $r_{t_j p_j}^{(j)}: t \times p \to \{0, 1\}, t, p \in \mathcal T$, $j$ is the order; column-stochastic matrix of pairwise types impacts (weights) $w \in \mathbb R^{N \times N}$; operator $W: r_{tp}^{(j)} \to \mathbb R^{ \| t \| \times \| p \| }$ that maps relation into corresponding column-stochastic adjacency matrix. If $r_{tp}$ is not defined for some $(t, p) \in \mathcal T^ 2$, then $w_{tp} = 0$. 

\begin{algorithm}[h]
    \caption{Vectorised Tensor SimRank for HSM}\label{tensorsimrank:vectorized}
 \KwData{$\mathcal T$ - classes, $\mathcal R$ - relations, $w$ - relation weights}
 \KwResult{ $\mathcal S = \{ s_t(a, b) \}_{t \in T}$ }
 \For{$t \in \mathcal T$}
 {
    $s_t^{(0)} = I$
 }
 k = 0 \\
 \Repeat{$ \sum_{t \in \mathcal T} \| s_t^{(k+1)} - s_t^{(k)} \| \leq \varepsilon $}
 {
  \For{$t \in \mathcal T$}
  {
   $s_t^{new} = 0$ \\
   \For{$\mathcal R \ni r: t \times p \mapsto \{ 1, 0 \}$}
   {
    $s_t^{new} = s_t^{new} + w_{tp} W(r_{tp}) s_p^{(k)} W(r_{pt})$
   }
    k = k + 1
  }
  \For{$t \in \mathcal T$}
  {
      $s_t^{(k+1)} = s_t^{new} - diag(s_t^{new}) + I$
  }
 }
\end{algorithm}

To achieve better results (see above)  on sparse relations we adopted the Low-Rank SimRank approximation \cite{DBLP:journals/corr/OseledetsO14} that uses Probabilistic Singular Value Decomposition \cite{halko2011finding} to perform fast approximate projections on low-rank matrix manifold at each step of the iterative process (Algorithm~\ref{tensorsimrank:lowrank}).

\begin{algorithm}[h]
    \caption{Low-rank Tensor SimRank for HSM}\label{tensorsimrank:lowrank}
 \KwData{$\mathcal T$ - classes, $\mathcal R$ - relations, $w$ - relation weights, $\{a_t\}$ - approximation ranks}
 \KwResult{ $\mathcal S = \{ s_t(a, b) \}_{t \in T}$ }
 \For{$t \in \mathcal T$}
 {
    $s_t^{(0)} = I$
 }
 k = 0 \\
 $u_t = 0$ \\
 $d_t = 0$ \\
 \Repeat{$ \sum_{t \in \mathcal T} \| s_t^{(k+1)} - s_t^{(k)} \| \leq \varepsilon $}
 {
  \For{$t \in \mathcal T$}
  {
   $s_t^{new} = 0$ \\
   \For{$\mathcal R \ni r: t \times p \mapsto \{ 1, 0 \}$}
   {
   \begin{multline*}
    s_t^{new} = s_t^{new} + w_{tp} ( W(r_{tp}) W(r_{pt}) + \\ + W(r_{tp}) u_p d_p u_p^T W(r_{pt}))
    \end{multline*}
   }
    k = k + 1
  }
  \For{$t \in \mathcal T$}
  {
  	  $s_t^{new} = s_t^{new} - T$
  	  $u_t, d_t = \text{ProbabilisticSVD}(s_t^{new}, a_t)$
      $s_t^{(k+1)} = s_t^{new} + I$
  }
 }
\end{algorithm}

The only difference with Algorithm~\ref{tensorsimrank:vectorized} is that on each step we perform probabilistic SVD decomposition of the matrix $S - I$, so that $S \approx I + UDU^T$, and project it onto the manifold of matrices of rank $a_t$.

\subsection{Convergence conditions}
Recall that  the classical SimRank can be computed as a solution of the equation:
 $$ S := W S W^T - \diag(W S W^T) + I.$$
Fixed-point iteration converges if $W$ is a column-stochastic matrix. In the vector form ($\vec(\cdot)$ operator maps an $n \times n$ matrix into a $n^2$ vector by taking column by column) that can be written as\footnote{$\vec(ABC) = (C^T \otimes A) vec(B)$}:
\[
[W \otimes W - I] \vec(S) - \vec(\diag(W S W^T)) + \vec(I) = 0,
\]
if matrix $W$ is stochastic, then $W \otimes W$ is stochastic too.

Tensor SimRank \eqref{eq:tansor_simrank_matrix_sum} computation can be equivalently written in the form:
\begin{equation} 
\label{eq:tensor_simrank_sol}
S := \sum_{\gamma} w_{\gamma} W_{\gamma} S W_{\gamma}^T - \diag(\sum_{\gamma} w_{\gamma} W_{\gamma} S W_{\gamma}^T) + I,\end{equation} 
or in the vectorized for 
$$ [\sum_{\gamma} w_{\gamma} W_{\gamma} \otimes W_{\gamma} - I] \vec(S) - \vec(\diag(\dots)) + \vec(I) = 0.$$
Moreover, SimRank is also commonly approximated by the solution of the discrete Lyapunov equation:
$$S = cW S W^T + (1-c)I, $$ 
 which can be generalized to the tensor case as
$$S = c \sum_{\gamma} w_{\gamma} W_{\gamma} S W_{\gamma}^T + (1-c)I, $$
and a fixed-point iteration converges \cite{bierkens2010estimate} if:
$$ \sum_{{\gamma}=1} w_{\gamma} \| W_{\gamma} \|_1^2 \le 1 \xLeftrightarrow[\text{stochastic}]{\| W_{\gamma} \|_1 = 1} \sum_{\gamma} w_{\gamma} \le 1.$$
We conjecture that fixed-point iterations for~(\ref{eq:tensor_simrank_sol}) converge if:
\begin{enumerate}
\item Each $W_{\gamma}$ is stochastic
\item $\sum_{\gamma} w_{\gamma}$ = 1
\end{enumerate}
In the simplest form (we have no preferences among relations and classes) it reduces to (relations weight):
$$ w_{tp} = \frac {1} { \sum_m \| \{ r^{(j)}_{tm} \in \mathcal R \} \| } .$$

\section{Computational experiment}
\subsection{Synthetic data: convergence test}
\begin{figure}[h]
\centering
\includegraphics[width=8.2cm]{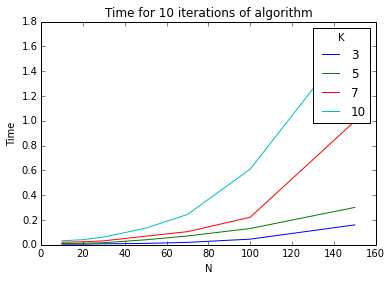}
\caption{Average time spent on 10 iterations of algorithm on randomly network with K components, N objects in each}\label{tensorsimrank:timefig}
\end{figure}

\begin{figure}[h]
\centering
\includegraphics[width=8.2cm]{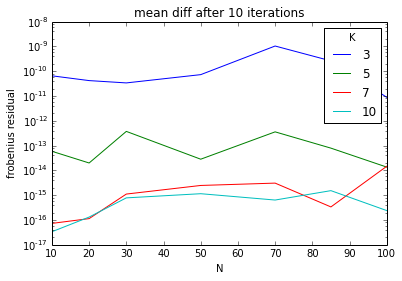}
\caption{Mean Frobenius residual after 10 iterations of algorithm as function of number of objects (N), K components}\label{tensorsimrank:meandiff}
\end{figure}

To test convergence conditions we conducted series of tests on randomly generated sparse networks with different number of classes: $K \in \{3, 5, 7, 10\}$ and with randomly chosen number of objects in each $N_{real} \in U_{[N/2; N]}$, $N \in \{10 \dots 100 \}$, full network of relation types (all possible types relations exists) with $2\min(N^i_{real}, N^j_{real})$ randomly chosen edges in each and default $w$ matrix (no priority). All generated networks successfully converged that illustrates that convergent sufficient conditions listed in previous section were adequate, see Figures~\ref{tensorsimrank:timefig},\ref{tensorsimrank:meandiff}.

\subsection{Synthetic data: similarity reconstruction}
 To determine if model is capable of similarity reconstruction we generated a tree graph from randomly distributed points on a plane and tested if model can reconstruct points spatial similarity basing only on their relations. 
 
\begin{figure}[h]
\centering
\includegraphics[width=8.2cm]{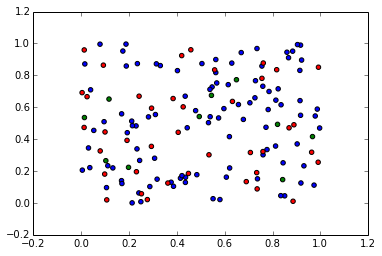}
\caption{Random points for graph generation: blue points -- zero level, red points -- first level, green point -- second level}\label{tensorsimrank:graph}
\end{figure}
 
\begin{algorithm}[h]
 \caption{Graph generation algorithm}
 \KwData{ $N$ - number of layers, $\{n_1, \dots, n_N\}$ - number of dots on each layer, $r$ - connection radius }
 \KwResult{ $\mathcal T$, $\mathcal R$ }
  \For{$k \in \{1..N\}$}
  {
  	$p^{(k)} \leftarrow$ generate $n_k$ point from $U_{[0;1]}^2$ \\
  	\If{$k > 0$}
  	{
  		$\mathcal R \leftarrow  r_k(p^{(k)}_i, p^{(k-1)}_j) $ if $\rho(p^{(k)}_i, p^{(k-1)}_j) < r $
  	}
  }
\end{algorithm}
On Figure~\ref{tensorsimrank:graph}  blue point represent 0-level point that are connected to 1-level point (red), that are connected to 2-level points (green).

We have measured the following similarity reconstruction $\hat S$ quality compared to real $S$ obtained from generated point coordinates:
$$ Q(S, \hat S) = \frac {\sum_a \sum_b \sum_c [S_{ab} < S_{ac} \ \textbf{and} \ \hat S_{ab} < \hat  S_{ac}]} {\sum_i \sum_j \sum_k 1}$$

that actually shows how many "$a$ is closer to $c$ then to $b$" relations were preserved.

 \begin{figure}
\centering
\includegraphics[width=8.2cm]{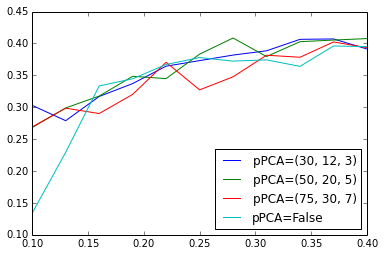}
\caption{The value of $Q(S, \hat S)$ as a function of $r$}\label{tensorsimrank:randomgraph}
\end{figure}

From Figure~\ref{tensorsimrank:randomgraph} one can see that at level $r \approx 0.3$ model gets saturated, but at the level $r \approx 0.15$ models that use low-rank version of Tensor SimRank perform way better than the "pure" algorithm. The numbers in the brackets denote the dimensionality of the matrix space into which the similarity matrices were projected on each step (rank of approximation).
\subsection{Book-Crossing Dataset test}
The model was run on subsample from the Book-Crossing Dataset \cite{ziegler2005improving}. We have extracted only those authors who had highest (top100) number of books in the collection. The final network had the following structure:
$$\mathcal T = \{\text{Book}, \text{Author}, \text{Year}, \text{Publisher} \}$$
$$\mathcal R = \{\text{isAuthorOf}(\cdot,\cdot), \text{publishedBy}(\cdot,\cdot), \text{publishedIn}(\cdot,\cdot) \}$$
$$ \# Book = 3625, \#Author = 99,$$
$$ \#Year = 65, \#Publisher = 554$$

\begin{figure}
\centering
\includegraphics[width=8.2cm]{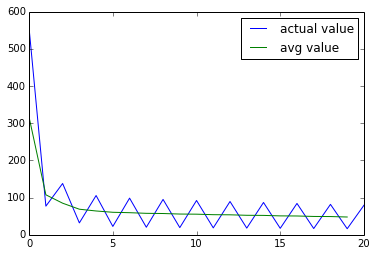}\label{tensorsimrank:books}
\caption{Monotonic reduction in the residual}
\end{figure}

Model convergence is shown on Figure~\ref{tensorsimrank:books}, where successfull convergence to the best possible low-rank approximation can be seen. The similarity structure is clearly visible on Year similarity matrix heatmap (Figure~\eqref{tensorsimrank:books}). We expect diagonal dominance as soon as temporarily close years should be more or less similar in terms of authors and publishers characteristic of that period. Tables~1 and 2 are examples of "closest book" requests, we want to notice that no NLP-preprocessing was conducted, nevertheless model treated books from same storybook as similar basing on author/publisher/year similarities.

\begin{figure}
\centering
\includegraphics[width=6cm]{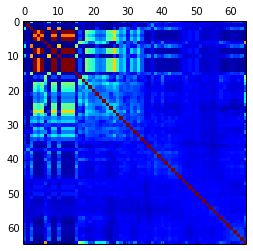}
\caption{Year similarity matrix}
\end{figure}

\begin{table}
\centering
\caption{Books closest to "Psychic Sisters"}
\begin{tabular}{|c|} \hline
\textbf{Psychic Sisters} \\ \textbf{(Sweet Valley Twins and Friends, No 70) } \\ \hline
The Love Potion \\ (Sweet Valley Twins and Friends, No 72) \\ \hline
The Curse of the Ruby Necklace \\ (Sweet Valley Twins and Friends Super, No 5) \\ \hline
She's Not What She Seems \\ (Sweet Valley High No. 92) \\ \hline
Are We in Love? \\ (Sweet Valley High,No 94) \\ \hline
Don't Go Home With John \\ (Sweet Valley High No. 90) \\ \hline
In Love With a Prince \\ (Sweet Valley High, No 91) \\ \hline
\end{tabular}
\end{table}

\begin{table}
\centering
\caption{Books closest to "The Girl Who Loved Tom Gordon"}
\begin{tabular}{|c|} \hline
\textbf{The Girl Who Loved Tom Gordon} \\ \hline
Hearts In Atlantis (All You Want to Know) \\ \hline
Blood And Smoke \\ \hline
Blood And Smoke Cd \\ \hline
Atlantis. \\ \hline
The Body (Penguin Readers: Level 5) \\ \hline
Storm of the Century \\ \hline
\end{tabular}
\end{table}

\section{Discussion and further work}
Proposed model can be used in various problem areas where most of the information is available in the form of relations between entities rather than features of individual entities and no trivial vector representation of those entities can be induced. 
One can use the vector representation
$$[s_t]_{ij} = \delta_{ij} + [u_t]_{ik} [d_t]_{kl} [u_t]_{lj} ,$$
to embed the notion of relations into classical machine learning algorithms. Also, the proposed model can be used for relation generalisation, that might give interesting results since we work on heterogeneous graphs.

Further model improvements might also include treating relations as objects too (probably, via heterogeneous hypergraphs) and defining similarity matrix on relations.

\section{Conclusion}
This paper proposes the generalization of SimRank for heterogeneous networks and a method for its computation that exploits the fact that the resulting similarity matrix is block-diagonal, thus its components might be computed in an iterative fashion. The convergence conditions are proposed and successfully tested. Few perspective application areas are suggested.
\bibliography{bib}
\end{document}